\documentclass[final]{cvpr}

\usepackage{times}
\usepackage{epsfig}
\usepackage{graphicx}
\usepackage{amsmath}
\usepackage{amssymb}
\usepackage{xspace}
\usepackage{booktabs}
\usepackage{soul}

\usepackage[pagebackref=true,breaklinks=true,colorlinks,bookmarks=false]{hyperref}

\def\cvprPaperID{****} 
\def\confYear{CVPR 2021}

\newcommand{\MPG}{\textit{MPG}\xspace}
\newcommand{\SLE}{\textit{SLE}\xspace}
\newcommand{\CR}{\textit{CR}\xspace}
\newcommand{\pizza}{\textbf{Pizza10}\xspace}

\DeclareMathOperator{\x}{\mathbf{x}}
\DeclareMathOperator{\y}{\mathbf{y}}
\DeclareMathOperator{\z}{\mathbf{z}}
\DeclareMathOperator{\te}{\mathbf{t}}
\DeclareMathOperator{\w}{\mathbf{w}}

\begin{document}

\title{MPG: A Multi-ingredient Pizza Image Generator with Conditional StyleGANs}  

\author{
Fangda Han\\
Rutgers University\\
Piscataway, NJ, USA\\
{\tt\small fh199@cs.rutgers.edu}
\and
Guoyao Hao\\
Rutgers University\\
Piscataway, NJ, USA\\
{\tt\small gh343@scarletmail.rutgers.edu}
\and
Ricardo Guerrero\\
Samsung AI Center\\
Cambridge, UK\\
{\tt\small r.guerrero@samsung.com}
\and
Vladimir Pavlovic\\
Rutgers University\\
Piscataway, NJ, USA\\
{\tt\small vladimir@cs.rutgers.edu}
}

\maketitle
\thispagestyle{empty}

\begin{abstract}
    Multilabel conditional image generation is a challenging problem in computer vision.
    In this work we propose \textbf{Multi-ingredient Pizza Generator (\MPG)}, a conditional Generative Neural Network (GAN) framework for synthesizing multilabel images. 
    We design \MPG based on a state-of-the-art GAN structure called StyleGAN2, in which we develop a new conditioning technique by enforcing intermediate feature maps to learn scalewise label information. 
    Because of the complex nature of the multilabel image generation problem, we also regularize synthetic image by predicting the corresponding ingredients.
    To verify the efficacy of \MPG, we test it on \textbf{Pizza10}, which is a carefully annotated multi-ingredient pizza image dataset.
    \MPG can successfully generate photo-realist pizza images with desired ingredients. The framework can be easily extend to other multilabel image generation scenarios.
\end{abstract}

\section{Introduction}

\begin{table}[ht]
\footnotesize
    \centering
    \caption{Glossary of terms.}
    \label{tab:glossary}
    \begin{tabular}{cc}
        \toprule
        Term & Description \\
        \midrule
        \MPG & Multi-ingredient Pizza Generator\\
        \SLE & Scalewise Label Encoder\\
        \CR & Classification Regularizer\\
        \pizza & Proposed multi-ingredient annotated pizza dataset\\
        \bottomrule
    \end{tabular}
\end{table}

Generative Adversarial Networks~\cite{goodfellow2014generative} (GAN) are a type of generative models that can create high-fidelity images~\cite{karras2019style, karras2020training, brock2018large} by alternatively optimizing its two components: generator and discriminator. We exploit the capacity of GAN models for solving the problem of generating food images conditioned on multiple ingredients. Food images are naturally complex since ingredients usually have different colors and shapes; even the same ingredient can have different colors with varying methods of cooking (\eg pepperoni will become darker with longer cooking time or higher cooking temperature). Moreover, ingredients in food images are usually mixed together, and therefore, the model has to learn to arrange the ingredients correctly on an image. 

Previous works on food image generation either use images as input~\cite{ito2018food, papadopoulos19cvpr} or generate blurry images~\cite{han2020cookgan}. We propose a conditional generative adversarial network called Multi-ingredient Pizza Generator (\MPG) inspired by~\cite{karras2020analyzing}. \MPG first encodes the ingredients into 
different scale-specific embeddings, injected into the synthesis network together with a ``style noise'' to create images. These same ingredient embeddings are also injected into the discriminator to guide the discrimination process.
 
Our main contributions are: 
(1) We propose a novel multilabel conditional GAN framework that is based on StyleGAN2~\cite{karras2020analyzing}. This new framework incorporates a Scalewise Label Encoder (\SLE) and a Classification Regularizer (\CR) that guide the generator in the synthesis of the desired ingredients.
(2) We create a new dataset \pizza from pizzaGANdata~\cite{papadopoulos19cvpr}, by relabiling using a subset of the ingredient labels. After re-annotation, these labels are able to perform more accurate in the multilabel classification and image retrieval tasks. 

The reminder of this manuscript is organized as follows: \autoref{sec:related_works} introduces the related works including GAN, conditional GAN, food generation and multilabel image generation. 
\autoref{sec:methodology} presents our conditional StyleGAN framework \MPG as well as the Scalewise Label Encoder (\SLE) and Classification Regularizer (\CR). 
\autoref{sec:pizza10} describes our motivation to create \pizza dataset and illustrates the improvement from the dataset reframing process.
\autoref{sec:experiments} includes experiments that verify the effectiveness of our framework, an ablation study, and an assessment of generated images.
Finally, in \autoref{sec:conclusion} we summarize our conclusions.

\section{Related Works}
\label{sec:related_works}

\noindent \textbf{GANs}~\cite{goodfellow2014generative} were proposed in 2014 and originally trained as a two-player min-max game to estimate the true data distribution from samples. 
In the past few years, GANs have evolved rapidly, improving their ability to generate high-fidelity digital images of compact objects and natural scenes~\cite{miyato2018spectral, brock2018large, karras2020training}, as well as human faces~\cite{karras2019style, karras2020analyzing}.
As GAN training involves two neural networks competing with each other, it is well-known to suffer from instability during training, leading to failure to converge or mode collapse~\cite{mescheder2018training, salimans2016improved}. 
Techniques proposed to solve these problems include mini-batch normalization~\cite{salimans2016improved}, R1 regularization~\cite{drucker1992improving, ross2017improving}, Wasserstein loss~\cite{arjovsky2017wasserstein, gulrajani2017improved}, Spectral Normalization~\cite{miyato2018spectral}  adaptive augmentation~\cite{karras2020training}, \etc.

\vspace{0.2em}
\noindent \textbf{Conditional GAN}~\cite{mirza2014conditional} extends GAN by conditioning generation process with auxiliary information to control the appearance of generated images, \eg controlling handwriting digit in MNIST~\cite{mirza2014conditional}, natural object or scene category in ImageNet dataset~\cite{miyato2018cgans}, or the category of CIFAR10 dataset~\cite{karras2020training}.
The work proposed by \cite{oeldorf2019loganv2} is the closest to our proposed framework. There, the authors apply StyleGAN to the logo generation task.
Our task differs substantially from this as we seek to condition on more than a single label (\eg ingredients); thus, the structures and spatial relationships among labels are far more diverse.
Another line of research that uses conditional GANs, aims at conditioning directly with an input image, with the goal of changing its attributes~\cite{he2019attgan, choi2018stargan} or associated styles~\cite{isola2017image, zhu2017unpaired}. However, it is important to note that the task of our proposed work is more complicated, as we only have labels as the conditioning input, lacking the input image as a strong prior.
Finally, image GANs can be conditioned directly on natural language textual descriptions~\cite{zhang2018stackgan++, xu2018attngan, li2019controllable}. While, in principle, more general, these approaches in practice work well on objects such as birds or flowers that share similar structures within classes, but tend to fail on complex scene generation, like those in COCO Dataset.

\vspace{0.2em}
\noindent \textbf{Food Image Generation} 
 is usually accomplished using a conditional GAN to attain higher fidelity compared with other generative models like VAEs~\cite{kingma2013auto} or Flow~\cite{kingma2018glow, dinh2014nice, dinh2016density}.
\cite{sabinigan} concatenates the embedding from a pretrained text encoder with noise to generated text-based food image.
\cite{el2019gilt} also uses the embedding of a pretrained text encoder but applies the more advanced StackGAN2~\cite{zhang2018stackgan++} to generate food image.
\cite{han2020cookgan} also applies StackGAN2 but it regularizes the generated image to retrieve the corresponding text.
\cite{wang2019food} applies ProgressiveGAN~\cite{karras2017progressive} to unsupervised food image generation on four different datasets.
However, these works typically result in blurry, low-resolution images with missing details; in contrast, our framework can create images of high fidelity and resolution.
\cite{ito2018food, papadopoulos19cvpr} apply GANs to transform the food type or add/remove ingredients of a source image. This improves the quality of generated images but requires image-based conditioning; our framework uses labels as input, a significantly weaker form of bias.

\vspace{0.2em}
\noindent \textbf{Multilabel Image Generation} 
is a special case of text-based image synthesis. \cite{zhang2018stackgan++} uses a stacked multi-scale generator to synthesize images based on text embeddings, ~\cite{xu2018attngan} extends~\cite{zhang2018stackgan++} and synthesize details at different subregions of the image by paying attentions to the relevant words in the natural language description. \cite{qiao2019mirrorgan} extends \cite{xu2018attngan} by regularizing the generator and redescribing the corresponding text from the synthetic image. While free text is a potentially stronger signal, it is often corrupted by image-unrelated signals, such as the writing style. On the other hand, while image labels may appear more specific than free-text, the visual diversity of objects described by those labels, such as food ingredients, and the changes in appearance that can arise from interactions of those objects can make the multilabel image generation problem an equal if not greater challenge to that of the caption-to-image task. 

\section{Methodology}
\label{sec:methodology}

\begin{figure}[t]
    \centering
    \includegraphics[width=0.46\textwidth]{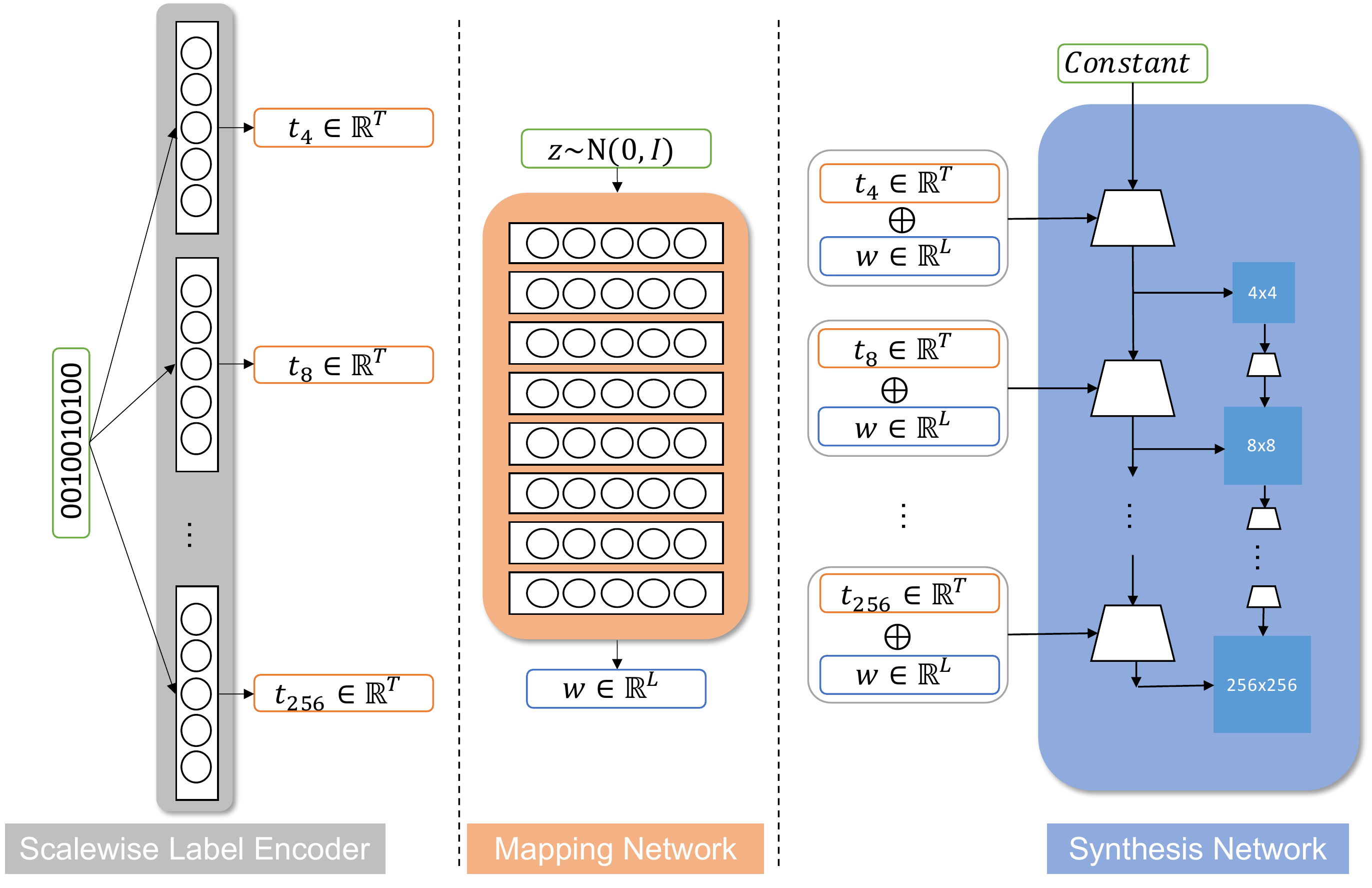}
    \caption{Generator components overview. \textbf{Left}: Scalewise Label Encoder (\SLE), each scale has its own layers for encoding. \textbf{Middle}: Mapping Network. \textbf{Right}: Synthesis Network, $\oplus$ means concatenation, notice images at different scales are conditioned with different label embedding}
    \label{fig:generator}
\end{figure}

\begin{figure}[ht]
    \centering
    \includegraphics[width=0.46\textwidth]{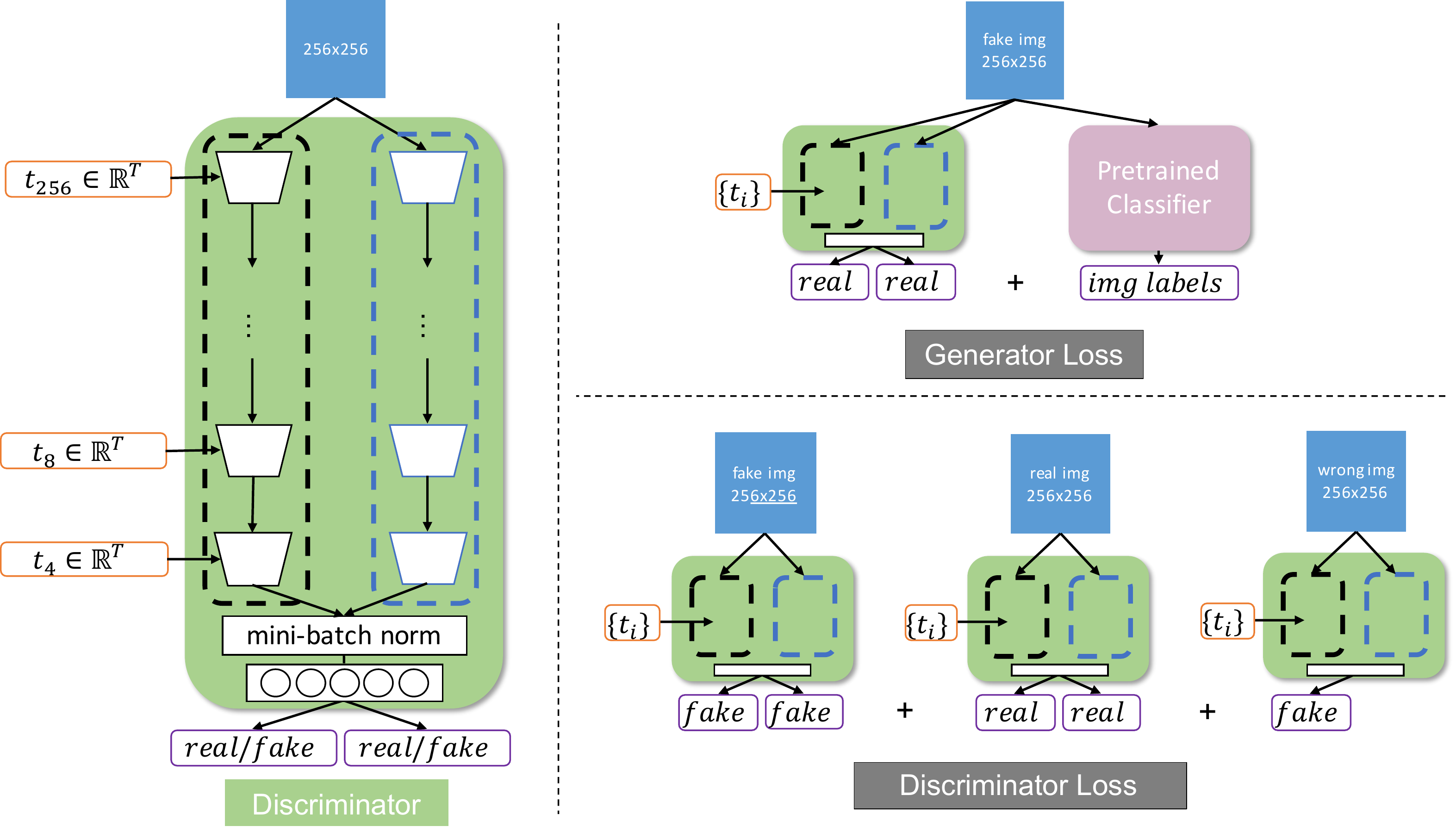}
    \caption{\textbf{Left}: Discriminator structure, which contains one branch for conditional output and another branch for unconditional output, notice the label embedding $\{\te_i\}$ are reversed to match different scales. \textbf{Top right}: Generator Loss, consisting discriminator loss plus Classification Regularizer (\CR) loss for fake image. \textbf{Bottom right}: Discriminator Loss, consisting three discriminator losses from fake, real and wrong images. Notice the discriminator is trained to distinguish between (txt, real img) and (txt, wrong img)}
    \label{fig:discriminator}
\end{figure}

\begin{figure}[ht]
    \centering
    \includegraphics[width=0.46\textwidth]{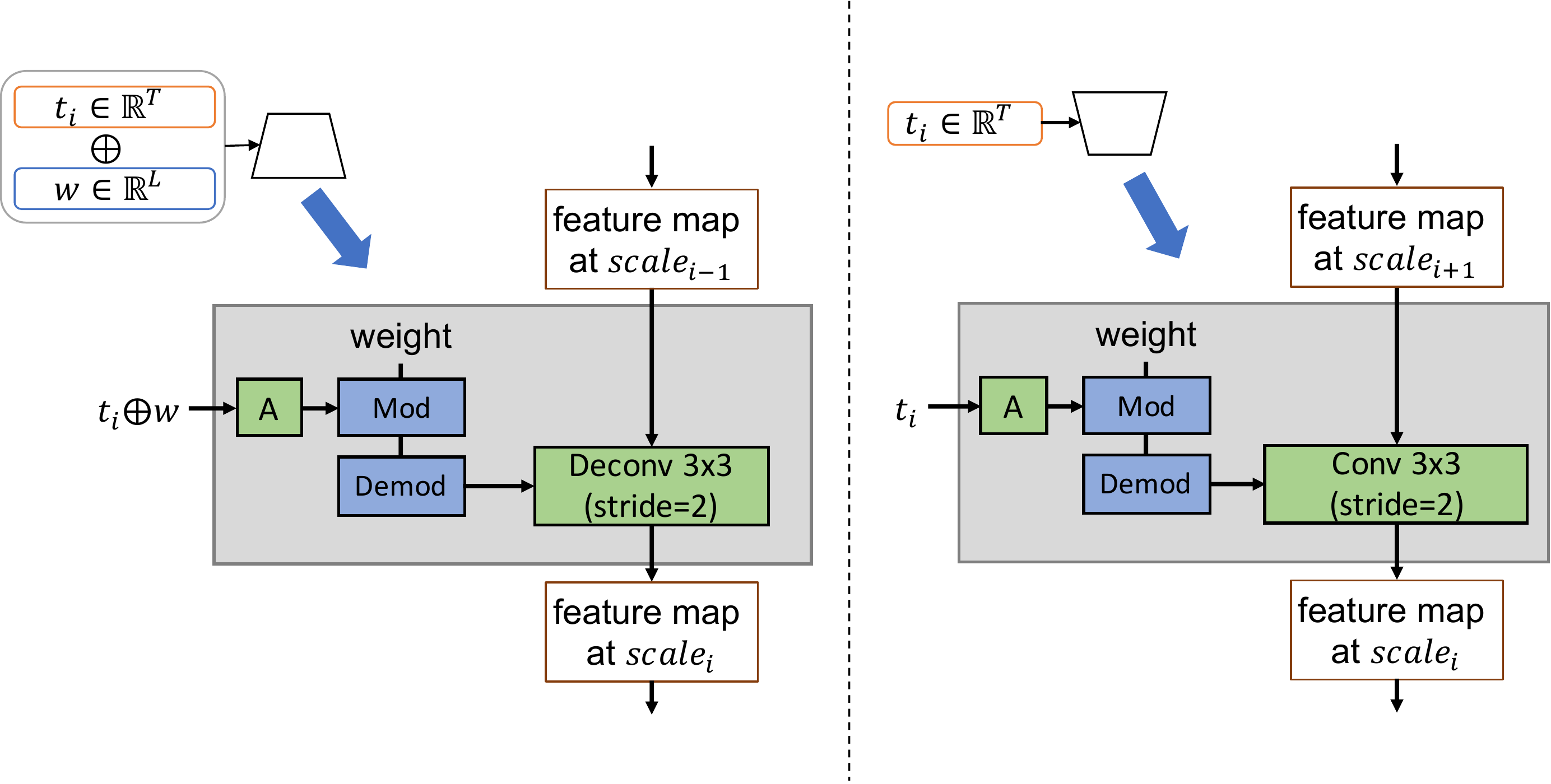}
    \caption{\textbf{Left}: Injection block at scale i in the generator (image courtesy \cite{karras2020analyzing}). \fbox{A} is a learned affine transformation, `weight' is the learned weight for the deconv layer which is modulated by $\te_i\oplus\w$ by \fbox{Mod} and \fbox{Demod}, refer \cite{karras2020analyzing} for more details. \textbf{Right}: Injection block at scale i in the discriminator. The same injection idea is applied to the discriminator to inject the label embedding at each scale, here conv layer is used instead of the deconv layer to downsample the feature map}
    \label{fig:inject_layer}
\end{figure}

To generate pizza images of high-fidelity, we start from StyleGAN2~\cite{karras2020analyzing}, one of the state-of-the-art GAN architecture. 
StyleGAN2 is composed of a generator (\autoref{fig:generator}) and a discriminator (\autoref{fig:discriminator} Left) just like the original GAN, but can generate high-fidelity human faces and objects of up to $1024^{2}$ resolution. 
There are several techniques involved to improve image quality. (1) The image is generated progressively. Lower-resolution image is generated first and then upsampled and combined with higher-resolution image. (2) There are two sources of noise as the input of the generator, one controls the style and the other controls the detail. The style noise ($z\in \mathcal{N}(0,I)$ in \autoref{fig:generator} Middle) and the detail noise (omitted in \autoref{fig:generator} because it is irrelevant to our work) are injected into the generator at each scale. The style noise is injected by first passing through a \textit{mapping network} (\autoref{fig:generator} Middle) and then transformed to be the weights of the convolutional layers in the generator, therefore achieving a strong control of the image, while the detail noise is injected directly by adding into the output feature map of each convolutional layer as a perturbation. (3) The discriminator is trained with mini-batch normalization~\cite{salimans2016improved} to avoid mode collapse of the generator. (4) R1 regularization~\cite{ross2017improving} is applied to improve the robustness of the network.

StyleGAN2 alone has a limited way of controlling the output image. The authors show that changing the style noise $\z$ will result in a shift in human face features (\eg face outline, skin color, gender \etc). However, given the style noise's random nature, it is impossible to determine or control specific features in the generated image. In this paper we adjust StyleGAN2 and create Multi-ingredient Pizza Generator (\MPG), making it possible to generate pizza images with specific, controllable ingredients and combinations thereof.

Formally, given a binary vector $\x$ indicating the ingredient list (\autoref{fig:generator} Left), we seek to generate a pizza image $\y$ that contains the desired ingredients. We propose to use a model endowed with Scalewise Label Encoder, Classification Regularizer to solve this problem.

\subsection{Scalewise Label Encoder}
The generator in StyleGAN2 is injected with the style noise $\z$ at each scale. 
Inspired by this, we design \SLE (Scalewise Label Encoder in \autoref{fig:generator} Left) to condition the synthesis network on the binary vector $\x$. 
\SLE consists of a sequence of sub-encoders $\{Enc_i(\x)\}$ (one for each scale $i$) and takes $\x$ as input to output label embedding $\{\te_i\}$ (also one for each scale $i$). Formally, if the generated image $\tilde{\y}$ is of size $256^2$, then we get scalewise label embedding $\{\te_i\}$
\begin{align}
    \{\te_i\} = SLE(x) = \{Enc_i(\x)\} \label{eq:sle},
\end{align}
where $i \in \{4,8,16,32,64,128,256\}$. $Enc_i$ is the encoder at $i-th$ scale. We implement $Enc_i$ by multi-layer perceptron. The mapping network $f$ is kept to provide more diversity to the output image. The label embedding $\{\te_i\}$ is then concatenated with the output from the mapping network together as the input of the synthesis network $S$. We inject $\te_i\oplus\w$ the same way as in StyleGAN2 (\autoref{fig:inject_layer} Left) and generate the output image $\tilde{\y}$
\begin{align}
    \w &= f(\z) \\
    \tilde{\y} &= S\left(\{\w, \te_i\} \right). \label{eq:syntheis_network}
\end{align}

\autoref{eq:sle} to \autoref{eq:syntheis_network} can be further simplified as
\begin{align}
    \tilde{\y} = G(\x, \z),
\end{align}
where G is the generator composed of $\{SLE, f, S\}$.

\SLE outputs are applied to the discriminator as well, since the discriminator is composed of a sequence of convolutional layers that downsample the input image from high resolution to low resolution, we reverse $\{\te_i\}$ and inject each $\te_i$ \textbf{without} concatenating $\w$ at each scale the same way as the generator (\autoref{fig:inject_layer} Right). However, during experiments we notice replacing the StyleGAN2 discriminator with the conditional version directly will lead to degeneration in image quality. Inspired by StackGAN2~\cite{zhang2018stackgan++}, we extract the convolutional layers in StyleGAN2 discriminator and add it into the discriminator as the ``unconditional branch'' (blue dash box in \autoref{fig:discriminator} Left). 

Formally, the discriminator $D$ is combined with a conditional branch $f_c$, an unconditional branch $f_{uc}$, a mini-batch normalization operation $\sigma$ and a fully-connected layer $F$. Given the label embedding $\{\te_i\}$ and input image $\y$, the outputs can be represented as
\begin{align}
    s_c &= D\left(\{\te_i\}, \y\right) = F\left(\sigma\left( f_c(\{\te_i\}, \y) \right)\right) \\
    s_{uc} &= D\left( \y \right) = F\left(\sigma\left( f_{uc}(\y) \right)\right),
\end{align}
where $s_c$ and $s_{uc}$ are conditional output and unconditional output respectively, $i\in \{256, 128, 64, 32, 16, 8, 4\}$. Notice mini-batch normalization is used to correlate samples in a mini-batch in the discriminator to prevent model collapse~\cite{salimans2016improved}.

Now given a pair of ingredient list and image $(\x, \y)$, the loss function for the generator can be defined as
\begin{align}
    \max_{G} = D( \{\te_i\}, G(\x, \z) ) + \lambda_{uncond} D(G(\x, \z)),
\label{eq:g_loss_0}
\end{align}
and the loss for the discriminator is
\begin{align}
\begin{split}
    \min_{D} 
    &= D( \{\te_i\}, G(\x, \z) ) + \lambda_{uncond} D(G(\x, \z)) \\
    &- D( \{\te_i\}, \y ) - \lambda_{uncond} D(\y),
\end{split}
\label{eq:d_loss_0}
\end{align}
where $\{\te_i\}=SLE(x)$, $\z\in \mathcal{N}(0,I)$, and $\lambda_{uncond}$ is the weight for unconditional output of the discriminator. 

\subsection{Classification Regularizer}
To help the generator create image with desired ingredients, inspired by~\cite{odena2017conditional, han2020cookgan}, we pretrain a multilabel classifier $h$ on pairs of real image and its ingredient list $\{(\y, \x)\}$. Then during training the generator, we regularize the generated image $\tilde{\y}$ to predict the correct ingredients $\x$. The Classification Regularizer (\CR) can be formalized as
\begin{align}
    \tilde{\x} = h(\tilde{\y}),
\end{align}
this regularizer can be easily combined in the generator loss \autoref{eq:g_loss_0} as shown in top right of \autoref{fig:discriminator},
\begin{align}
\begin{split}
    \max_{G} 
    &= D( \{\te_i\}, G(\x, \z) ) + \lambda_{c} D(G(\x, \z)) \\
    &+ \lambda_{clf}BCE\left(\x, h(\tilde{\y})\right),
\end{split}
\label{eq:g_loss_final}
\end{align}
where $BCE$ is binary cross entropy loss and $\lambda_{clf}$ is the weight for the regularizer.


Except the contents mentioned above, R1 regularization \cite{drucker1992improving} is also kept in the discriminator to penalize the gradient flowed from the image to the output, which has been proved to be helpful in stablize GAN training process~\cite{ross2017improving}. The final discriminator loss is now
\begin{align}
\begin{split}
    \min_{D} 
    &= D( \{\te_i\}, G(\x, \z) ) + \lambda_{uncond} D(G(\x, \z)) \\
    &- D( \{\te_i\}, \y ) - \lambda_{uncond} D(\y) \\
    &+ \lambda_{r1} \left( \dfrac{\partial D(\x)}{\partial \x} + \dfrac{\partial D(\x, \{\te_i\})}{\partial \x} \right).
\end{split}
\label{eq:d_loss_final}
\end{align}

\autoref{eq:g_loss_final} and \autoref{eq:d_loss_final} summarize all components in the proposed framework \MPG. In the following sections, we will use \pizza dataset to verify the efficacy of \MPG.

\section{Pizza10 Dataset}
\label{sec:pizza10}

\begin{figure}[t]
    \centering
    \includegraphics[width=0.45\textwidth]{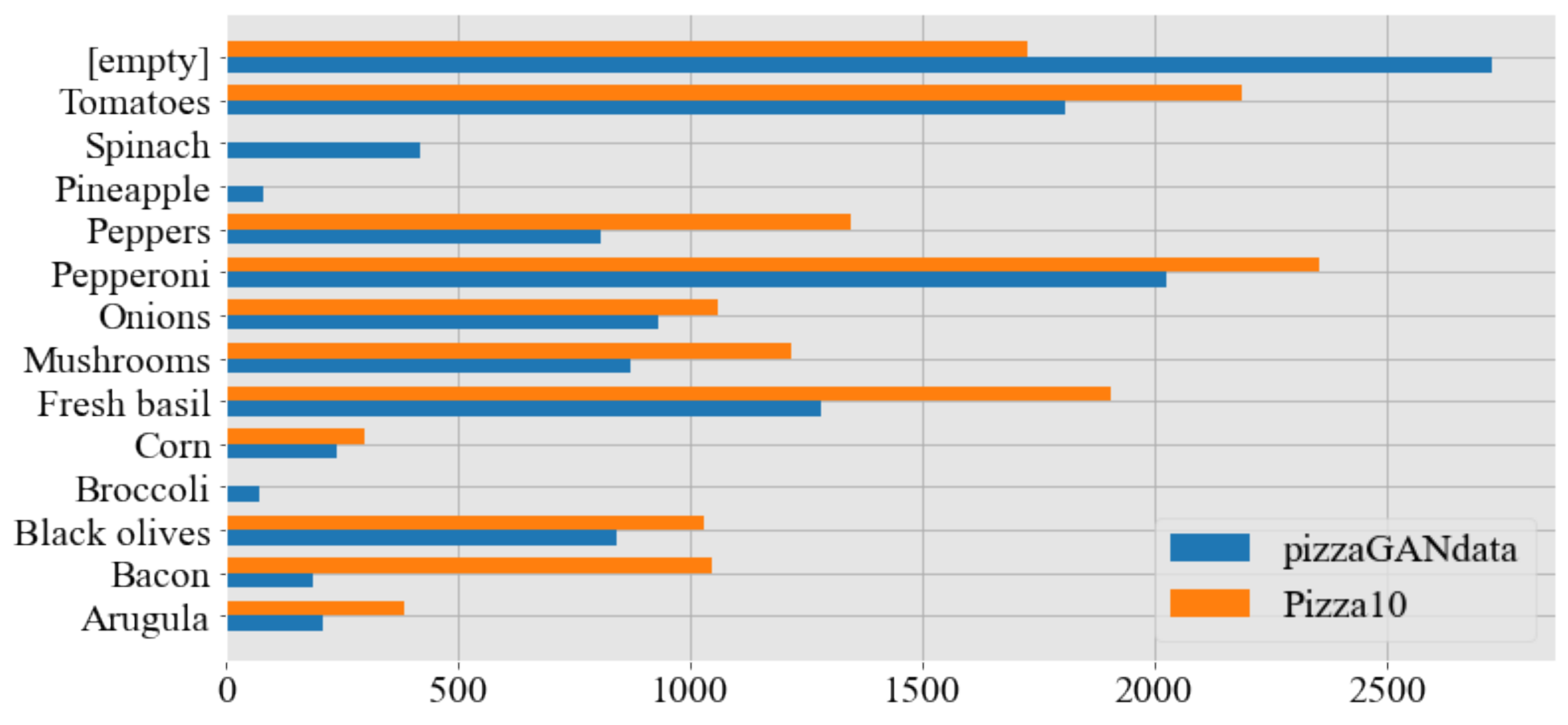}
    \caption{Ingredients distribution comparison between pizzaGANdata and the refined \pizza dataset, `[empty]' represents the number of samples with no ingredients labeled}
    \label{fig:ingredients_distribution}
\end{figure}

\begin{figure}[t]
    \centering
    \includegraphics[width=0.45\textwidth]{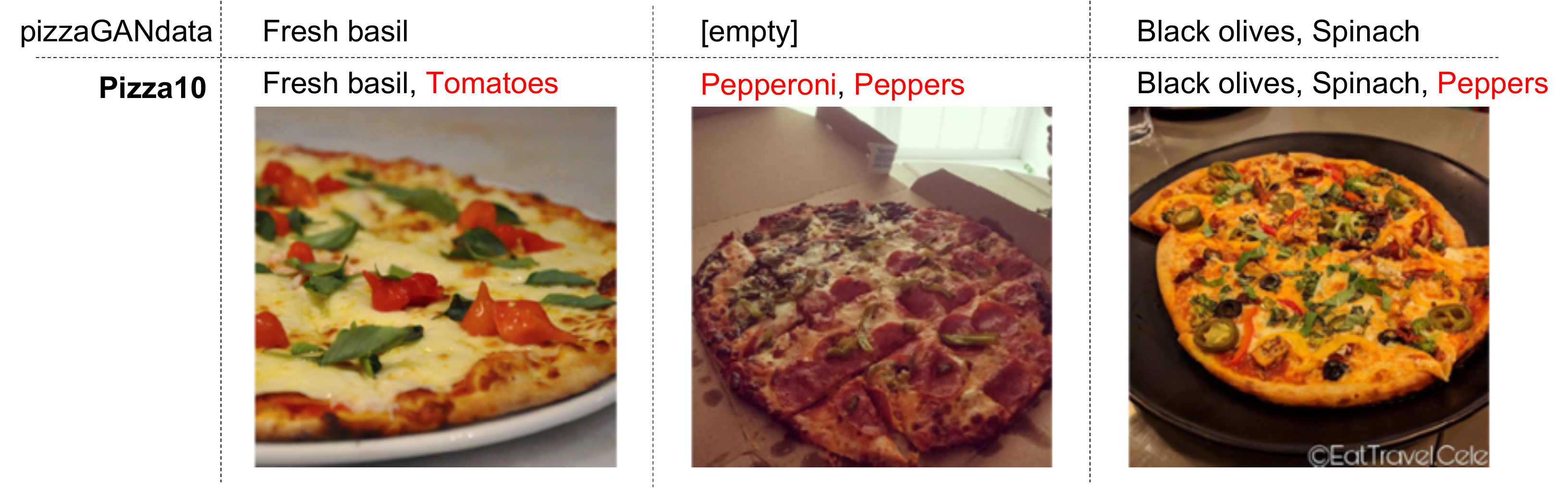}
    \caption{Illustration of improper labels in pizzaGANdata and the refined labels in \pizza dataset, the difference is highlighted in red (best viewed in color)}
    \label{fig:wrong_labeling}
\end{figure}

The performance of neural network models critically depends on the quality of the dataset used to train the models. 
We begin with the pizzaGANdata~\cite{papadopoulos19cvpr}, which contains $9213$ pizza images annotated with 13 ingredients, as shown in \autoref{fig:ingredients_distribution}. 
We manually inspected the labeling in this dataset and found a significant portion  of it to be mislabeled; we illustrate three instances in \autoref{fig:wrong_labeling}. 
To address this issue, using a custom-created collaborative web platform, we relabeled the dataset and removed three ingredients (\textit{Spinach, Pineapple, Broccoli}) with fewer number of instances (\textit{Corn} is kept since the distinct color and shape makes it a recognizable ingredient even with fewer samples than \textit{Spinach}). 
The refined dataset, \pizza, decreases the number of `empty samples' (pizza images with no ingredients labeled) by $40\%$ and increases the number of samples for each ingredient category as shown in \autoref{fig:ingredients_distribution}.
We verify the impact of the curated \pizza on two tasks:

\vspace{0.2em}
\noindent \textbf{Retrieval.} Given an ingredient list as query, we want to retrieve its associated pizza image, we solve this task by following~\cite{salvador2017learning}. The ingredient list is treated as a sentence and encoded into a vector of $\mathbb{R}^{768}$, the text encoder is two-layer-two-head transformer encoder~\cite{vaswani2017attention} built using HuggingFace\cite{wolf2019huggingface}, and we initialized the embedding from the pre-trained `bert-base-uncased' model on HuggingFace Model Hub. 
The image is also encoded into a vector of $\mathbb{R}^{768}$, the image encoder is initialized from ResNet50~\cite{he2016deep} with the final fully-connected layer changed to match the output dimension ($\mathbb{R}^{768}$). 
The loss is triplet loss with hard-sample mining~\cite{wang2019learning}.
We use $60\%$ data for training and leave $40\%$ for testing.
After training, we use the median rank (MedR)~\cite{salvador2017learning} to measure the performance; the retrieval range is $280$, roughly the number of distinct ingredient lists in both datasets. 
Computed as the median rank of the true
positives over all queries, a lower MedR ($\geq 1.0$) suggests better performance.

\vspace{0.2em}
\noindent \textbf{Multilabel Classification.} Given a pizza image, we seek to predict its ingredients. 
We initialized the classifier from ResNet50~\cite{he2016deep} with the final fully-connected layer changed to match the output dimension ($\mathbb{R}^{13}$ for pizzaGANdata and $\mathbb{R}^{10}$ for \pizza). 
The loss is a binary cross entropy loss. 
$80\%$ data are used in training while $20\%$ data are for testing.
After training, we use the mean average precision (mAP) as the metric.

\begin{figure}[t]
    \centering
    \includegraphics[width=0.45\textwidth]{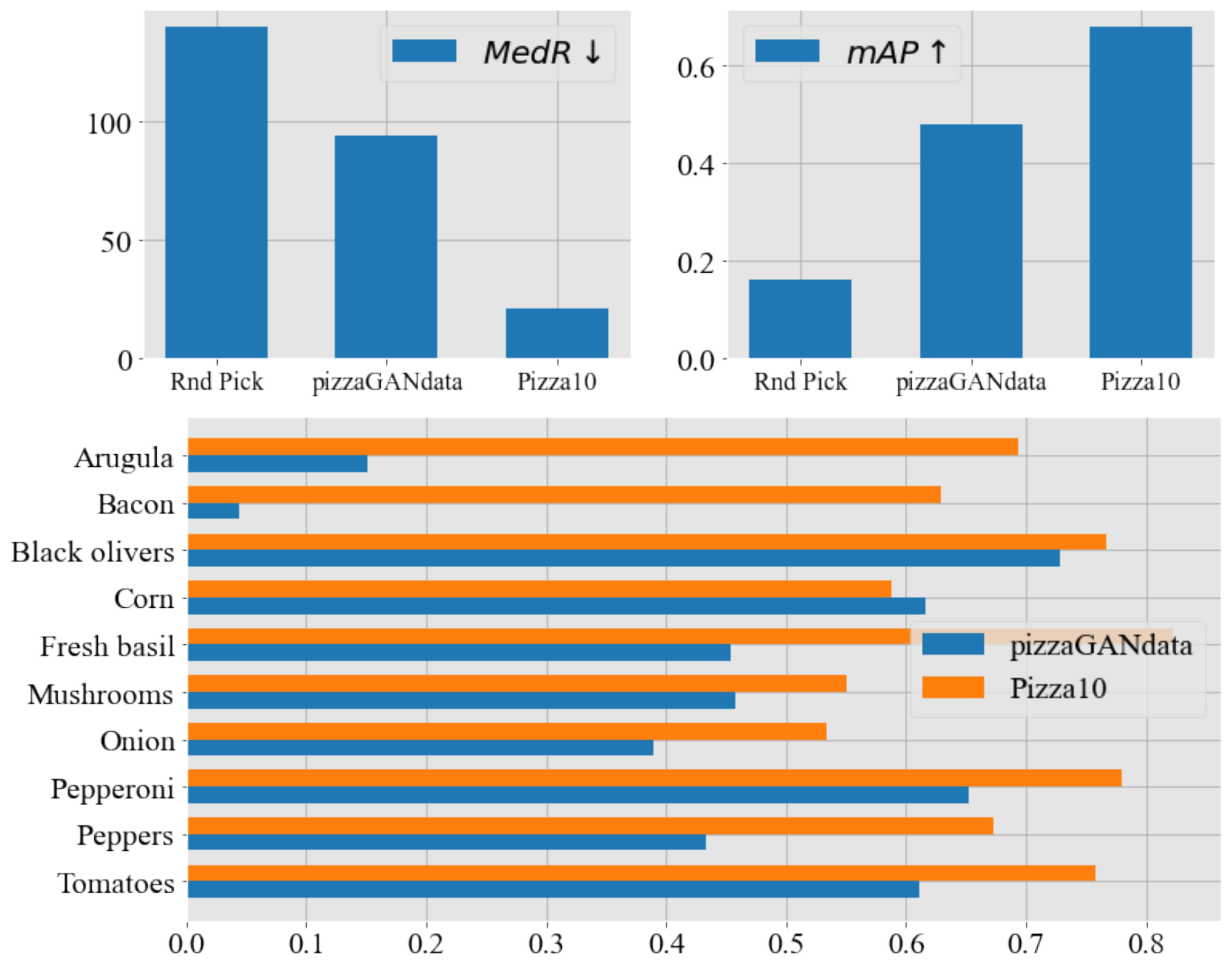}
    \caption{Performance comparison between pizzaGANdata and \pizza. Top left: MedR scores on retrieval task. Top right: mAP scores on multilabel classification task. Bottom: average precision for each ingredient in the classification task. `Rnd Pick' means the scores when we randomly pick answer. $\downarrow$ means the lower the better, $\uparrow$ means the higher the better}
    \label{fig:compare_dataset_performances}
\end{figure}

The results are shown in \autoref{fig:compare_dataset_performances}. The same model has a big jump in performance on the refined \pizza dataset in both tasks. Upon close inspection, the average precision (AP) for each ingredient increases on most ingredients except \textit{Corn} for the classifier trained on \pizza, therefore demonstrating the importance of careful curating.

\begin{figure*}[ht]
    \centering
    \includegraphics[width=0.86\textwidth]{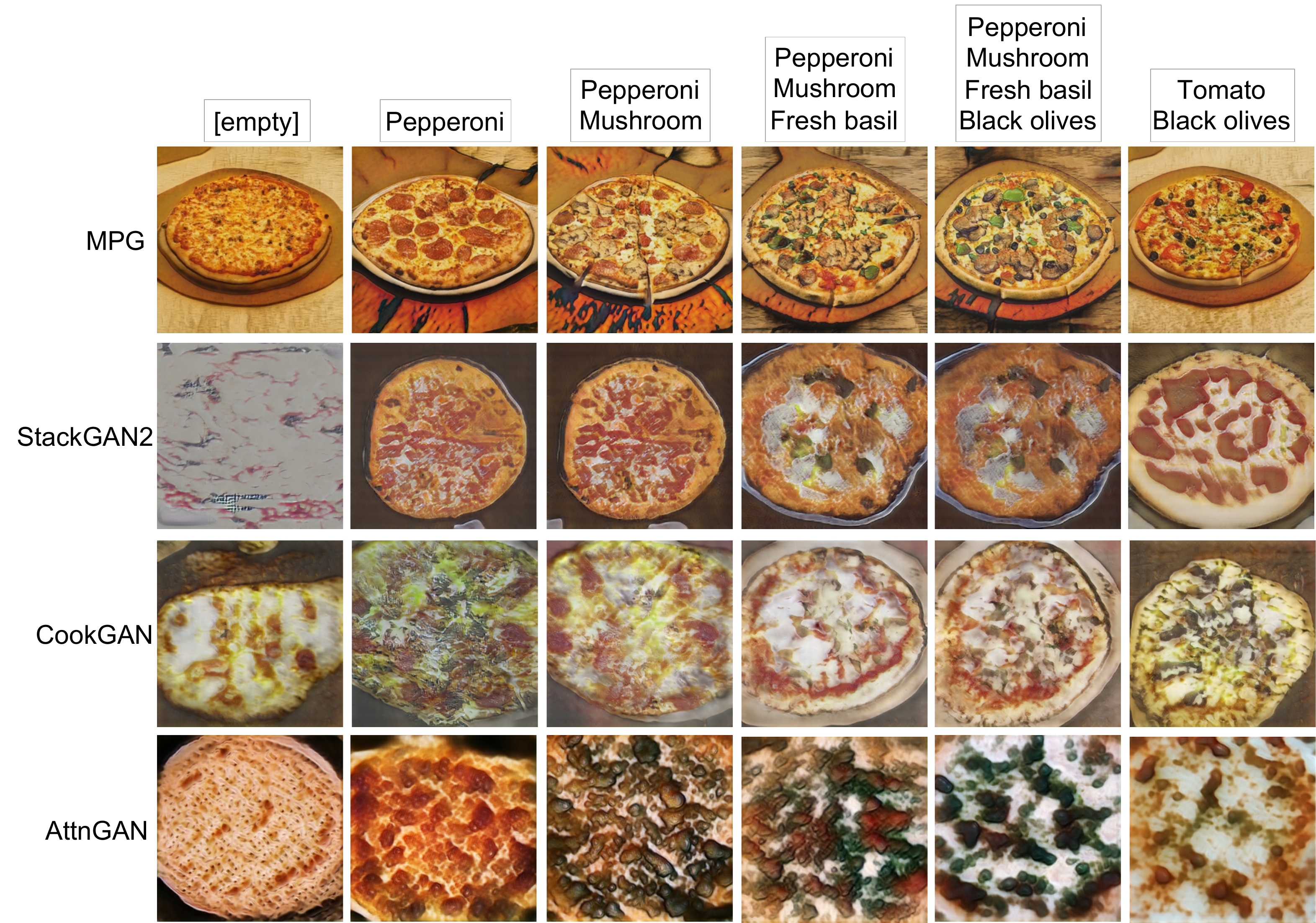}
    \caption{Qualitative comparison between our model \MPG and baselines}
    \label{fig:qualitative_comparison}
\end{figure*}

\section{Experiments}
\label{sec:experiments}
\noindent \textbf{Network Structures.} For the generator, \SLE is composed of a group of sub-encoders, each sub-encoder is a fully-connected layer with ReLU activation. The label embedding dimension $T=256$, the mapping network is the same as StyleGAN2 except we decrease the style noise dimension $L$ to $256$. The synthesized network is also the same as StyleGAN2 however the input at each scale is now different for matching the scalewise label embedding $\{\te_i\}$. For the discriminator, the unconditional branch, mini-batch normalization and fully-connected layer are the same as StyleGAN2, the conditional branch is modified from the injection block of the generator and performs the reversed operation (\eg injection and downsampling).

\vspace{0.2em}
\noindent \textbf{Hyper Parameters.} For the generator, $\lambda_{uncond}=1.0$. For the discriminator, $\lambda_{clf} = 1.0$, $\lambda_{r1} = 10$, R1 regularization is performed every 16 mini-batches to balance the training time. The learning rate for both the generator and discriminator is $0.002$. The mini-batch size is $24$ (need at least 24GB GPU memory).

\vspace{0.2em}
\noindent \textbf{Configurations.} The framework is built using PyTorch 1.7 and trained on four NVIDIA K80 GPUs (each has 12 GB memory), the training takes 6 days to converge for a duration of 2.4M pizza images. Note the number of images needed is much smaller than the StyleGAN model trained on LSUN BEDROOMS, CATS or CARS (see Appendix E in \cite{karras2019style}), we believe this is because our dataset is much smaller (\eg $9213$ images in \pizza vs.\ 3M in LSUN BEDROOMS).

\vspace{0.2em}
\noindent \textbf{Metrics.} We use FID~\cite{heusel2017gans} (on 10K images) to assess to the quality of the generated images, a lower FID indicates the generated image distribution is more close to the real image distribution. We use mean average precision~\cite{zhu2004recall} ($mAP$) to evaluate the conditioning on the desired ingredients. The multilabel classifier we used is trained on \pizza as shown in \autoref{sec:pizza10} and achieves $mAP=0.6804$.

\subsection{Comparison with Baselines}

We compare with three prior works on text-based image generation. \textbf{StackGAN2}~\cite{zhang2018stackgan++} first extracts text feature from a pretrained retrieval model and then forward through a stacked generator with separate discriminators working at different scales. \textbf{CookGAN}~\cite{han2020cookgan} uses similar structure as of StackGAN2 and adds a cycle-consistent loss which drives the generated images to retrieve its desired ingredients. \textbf{AttnGAN}~\cite{xu2018attngan} also shares StackGAN2-like structure, it uses pretrained text encoder to capture the importance of each word and leverages both word-level and sentence-level feature to guide the generation process.

\begin{table}[t]
    \centering
    \caption{Quantitative comparison of performances between baselines and the proposed multi-ingredient Pizza Generator (\MPG)}
    \label{tab:quantitative_comparison}
    \begin{tabular}{cccc}
        \toprule
        Models & Image Size & FID$\downarrow$ & mAP$\uparrow$ \\
        \midrule
        StackGAN2~\cite{zhang2018stackgan++} & $256^2$ & $63.51$ & $0.3219$ \\
        CookGAN~\cite{han2020cookgan} & $256^2$ & $45.64$ & $0.2896$ \\
        AttnGAN~\cite{xu2018attngan} & $256^2$ & $74.47$ & $0.5729$ \\
        \MPG & $256^2$ & $\mathbf{8.43}$ & $\mathbf{0.9816}$ \\
        \bottomrule
    \end{tabular}
    \vspace{-1em}
\end{table}

\begin{figure*}[ht]
    \centering
    \includegraphics[width=0.86\textwidth]{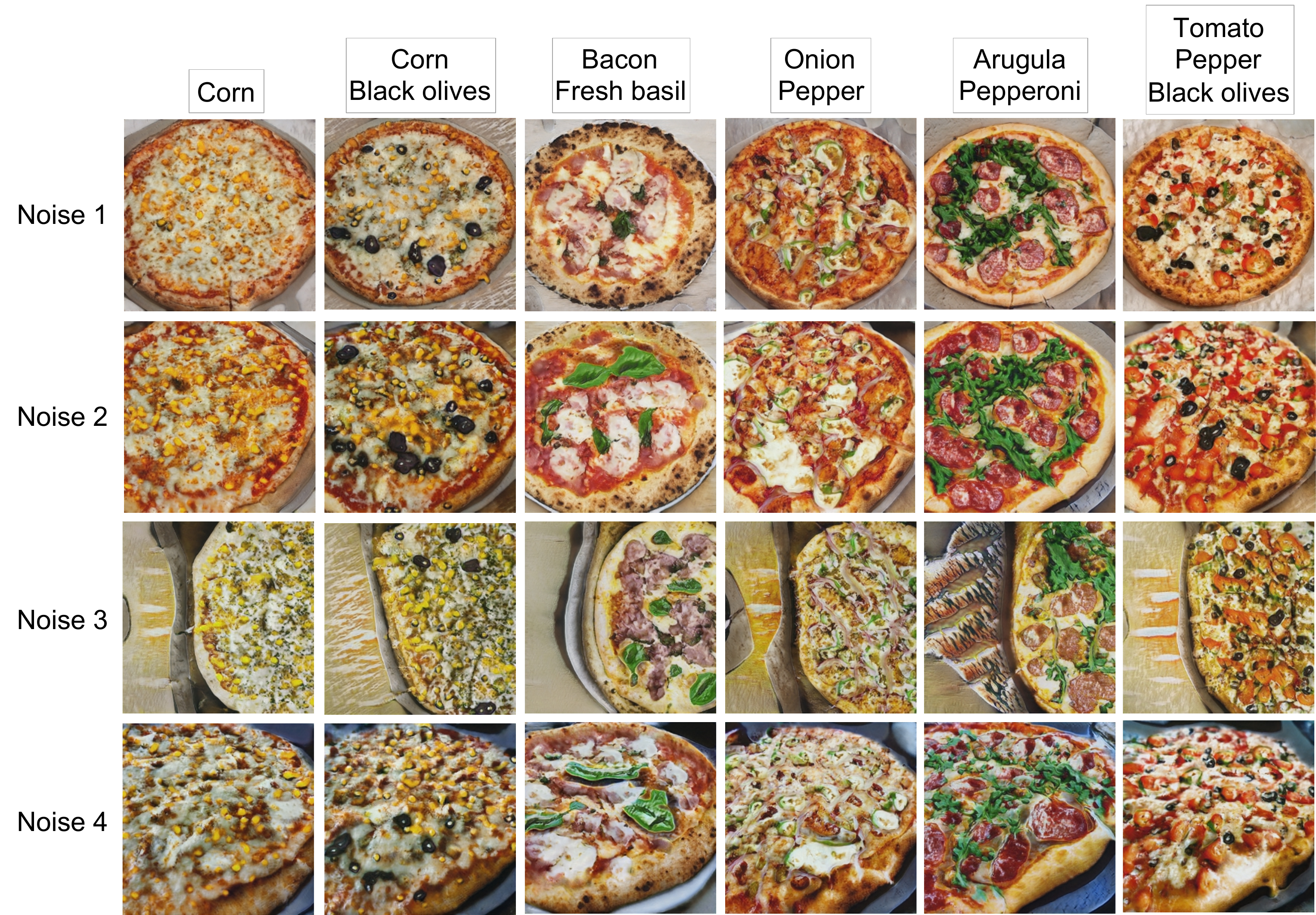}
    \caption{Images generated from different combinations of ingredient list and style noise. Images in the same row are generated with identical style noise.}
    \label{fig:independence}
\end{figure*}

\autoref{tab:quantitative_comparison} compares the performance between our model \MPG and the baselines, we can see \MPG has a huge leap forward compared with previous work both in terms of FID and mAP, which means \MPG not only generates realistic images with diversity but also improves the conditioning on the desired ingredients. 

One thing needs to mention is FID has a lower bound which can be estimated from real images, say if we randomly split the real images of \pizza into two image sets (about 5K images each set) and compute the FID between the two sets, we get FID$ \approx 8.20$, which means our model still has space to improve. Another thing to note is that mAP is much higher on the generated images from \MPG even compared with that on real images ($0.6804$), this sounds unreasonable but we believe since our model is trained with limited data, the images generated is good at capturing the main characteristic of each ingredient and incapable of understanding all possible diversities, while real images could be even harder for the classifier since they contain more variety.

\begin{table}[t]
    \centering
    \caption{Quantitative comparison of performances between \MPG and its counterparts with missing components}
    \label{tab:ablation_study}
    \begin{tabular}{ccc}
        \toprule
        Models & FID$\downarrow$ & mAP$\uparrow$ \\
        \midrule
        \MPG-mapping& $9.47$ & $0.9799$ \\
        \MPG-\SLE & $9.29$ & $0.9187$ \\
        \MPG-\SLE* & $10.03$ & $0.9276$ \\
        \MPG-\CR  & $9.85$ & $0.4712$ \\
        \MPG-uncond  & $22.26$ & $\textbf{0.9856}$ \\
        \MPG & $\textbf{8.43}$ & $0.9816$ \\
        \bottomrule
    \end{tabular}
    \vspace{-2em}
\end{table}

\autoref{fig:qualitative_comparison} shows a few samples from our model \MPG as well as the baselines conditioned on certain ingredient lists. Images generated by \MPG are more appealing and contain the desired ingredients compared with baselines. Notice how \MPG can effectively adds \textit{Pepperoni, Mushroom, Fresh basil and Black olives} one after another into the generated images.

\subsection{Ablation Study}
To effectively verify the design of \MPG, we bring out the following experiments. \textbf{\MPG-mapping} removes the mapping network $f$ and directly concatenates scalewise label embedding $\{\te_i\}$ with style noise $z$ as the input of the synthesis network. \textbf{\MPG-\SLE} removes Scalewise Label Encoder and use the same label embedding for all resolutions. \textbf{\MPG-\SLE*} also removes \SLE and uses the same label embedding for all resolutions, but the label embedding is concatenated with the style noise $\z$ before passing through the mapping network, instead of concatenating with $w$ after passing the mapping network. \textbf{\MPG-\CR} removes Classification Regularizer. \textbf{\MPG-uncond} removes the unconditional branch in the discriminator.

\autoref{tab:ablation_study} shows the performances of all models, we can observe that mapping network (\textbf{\MPG-mapping}) and unconditional branch in discriminator (\textbf{\MPG-uncond}) have impact on FID. Classification Regularizer (\textbf{\MPG-\CR}) mainly affects mAP. And \textbf{\SLE} influences both FID and mAP.

\subsection{Assessment of Generated Images}

We provide additional analyses of images created by \MPG. We first qualitatively study the independence between text embedding and style noise, then we assess the smoothness of the text and the style noise spaces.

\noindent \textbf{Independence.} To analyze the independence between text embedding and style noise, we fix the style noise while generating images with different ingredient lists. The result is shown in \autoref{fig:independence}. Taking the camera position of the images from Noise 1 as the anchor, Noise 2 provides a zoom-in top view, Noise 3 translates the camera position along x-axis, and Noise 4 offers a scaled side view of the pizzas. The likeness of view points in each row and the consistency of ingredients in each column demonstrate the independence between text embedding and style noise.

\begin{figure}[ht]
    \centering
    \includegraphics[width=0.47\textwidth]{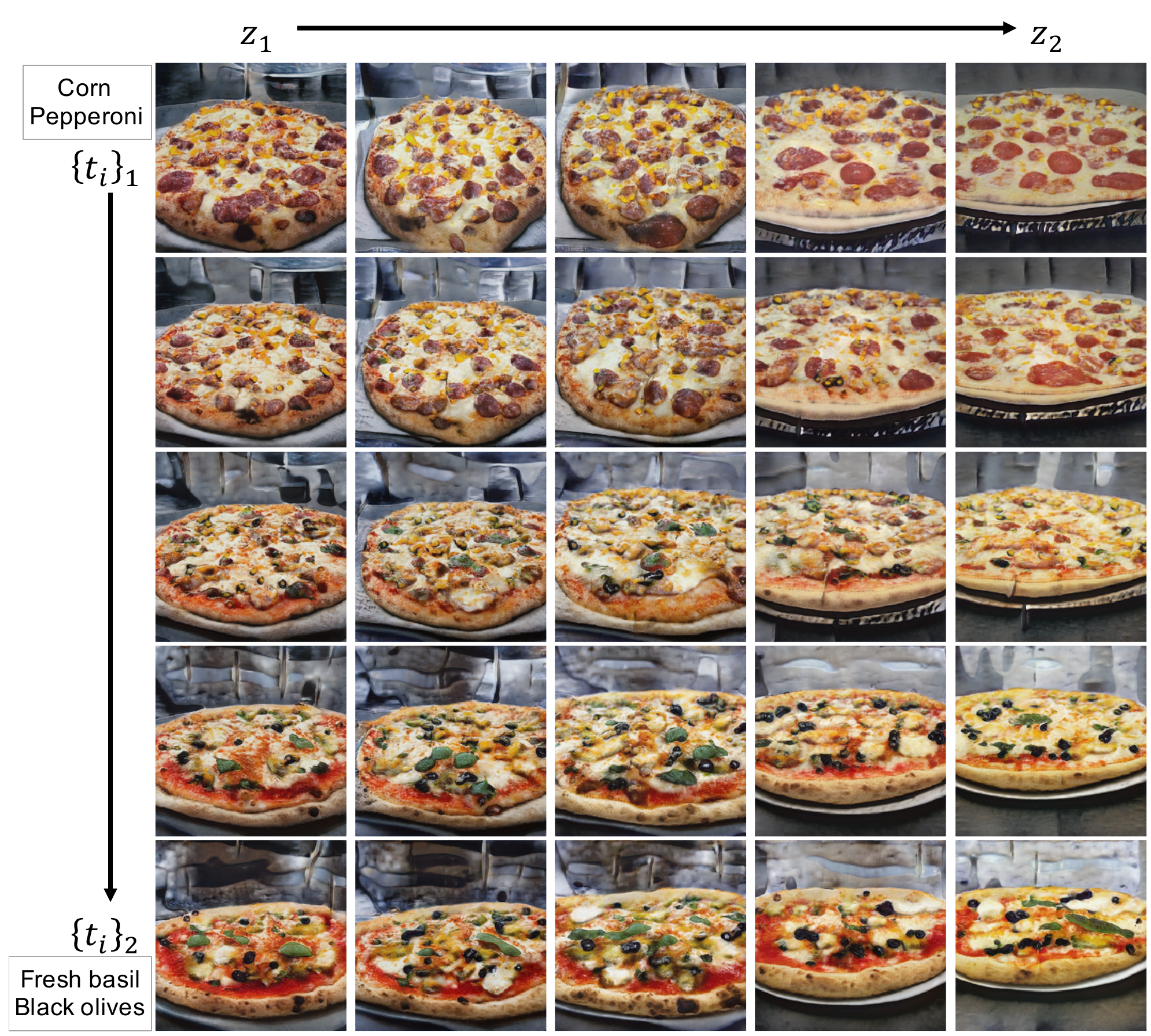}
    \caption{Illustration of traversing through the text embedding space and the style noise space. Images in each row are generated with the same text embedding (interpolated between $\{\te_i\}_1$ and $\{\te_i\}_2$), while images in each column are synthesized with the same style noise (interpolated between $\z_1$ and $\z_2$).}
    \label{fig:interpolation}
\end{figure}

\noindent \textbf{Smoothness.} We verify the smoothness of the learned embedding space and the style noise space by traversing~\cite{goodfellow2014generative}. In \autoref{fig:interpolation}, we randomly select two ingredient lists $\x_1$ and $\x_2$, forward them through \SLE and obtain the corresponding text embeddings $\{\te_i\}_1$ and $\{\te_i\}_2$; we then sample two style noise $\z_1$ and $\z_2$. The traversing is conducted by linear interpolation (\eg $\{\te_i\}_1 \rightarrow \{\te_i\}_2$ and $\z_1 \rightarrow \z_2$). 

We can observe the image generated with (\textit{Corn}, \textit{Pepperoni}) in the top left corner gradually changes to the images containing (\textit{Fresh basil}, \textit{Black olives}) along the vertical axis, while the view angle and coloring progressively update along the horizontal axis, from $\z_1$ to $\z_2$.  This demonstrates the ability of \MPG to learn, then recreate independent and smooth major factors of variation in this complex image space.

\section{Conclusion}
\label{sec:conclusion}

We propose \MPG, a Multi-ingredient Pizza Generator image synthesis framework, supported by a new \pizza dataset, that can effectively learn to create photo-realistic pizza images from combinations of specified ingredients, as well as manipulate the view point by independently changing the style noise. While extremely effective in visual appearance, the model remains computationally expensive to train and with some loss of detail at fine resolutions, the problems we aim to investigated in future research.

{\small
\bibliographystyle{ieee_fullname}
\bibliography{egbib}
}

\newpage
\appendix
\section{Video}
\begin{figure*}
    \centering
    \includegraphics[width=0.9\textwidth]{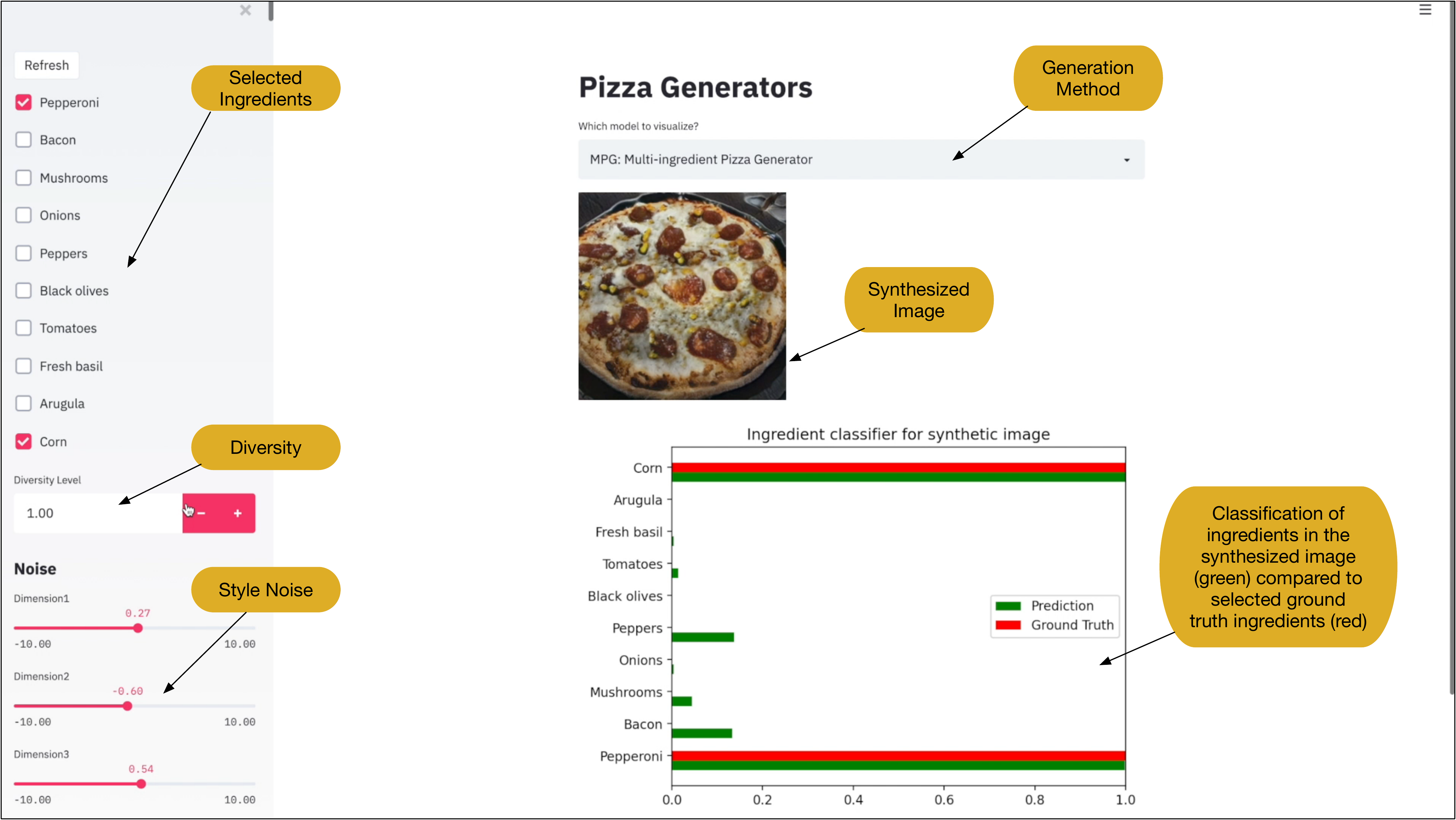}
    \caption{Web app for synthesis of photo-realistic pizza images using the proposed MPG framework.  Oval legend blocks explain the main elements used in the demo video.}
    \label{fig:video_explain}
\end{figure*}

In the video\footnote{https://youtu.be/x3XKXMd1oC8}, we show two examples to demonstrate the main features of our proposed Multi-ingredients Pizza Image Generation (MPG) framework. The first example is an interactive web app we built that shows the images generated from arbitrary selected ingredients and style noise.  See \autoref{fig:video_explain} for details about the app layout.

\begin{figure*}
    \centering
    \includegraphics[width=0.9\textwidth]{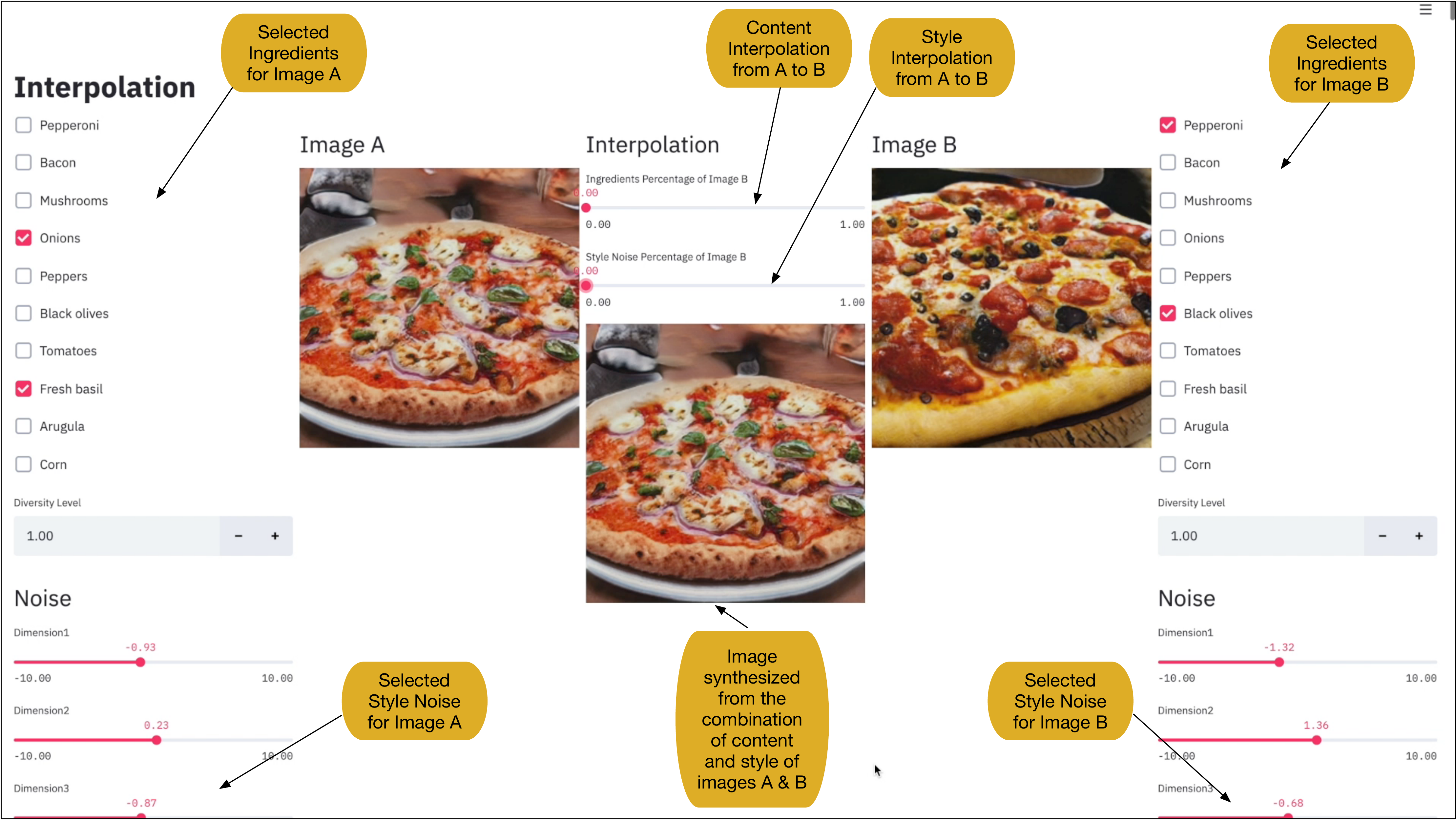}
    \caption{Web app for synthesis of photo-realistic pizza images using the interpolation approach. Image in the middle is created by interpolating the content and the style of used to create the images A and B. Oval legend blocks explain the main elements used in the demo video.}
    \label{fig:video_explain_interp}
\end{figure*}

The second example, showcased in \autoref{fig:video_explain_interp}, is another interactive web app we designed to explore the smoothness of the learned ingredient representation space. We encourage the reviewers to watch the included video.

\section{Additional Visual Comparisons}
\begin{figure*}[ht]
    \centering
    \includegraphics[width=\textwidth]{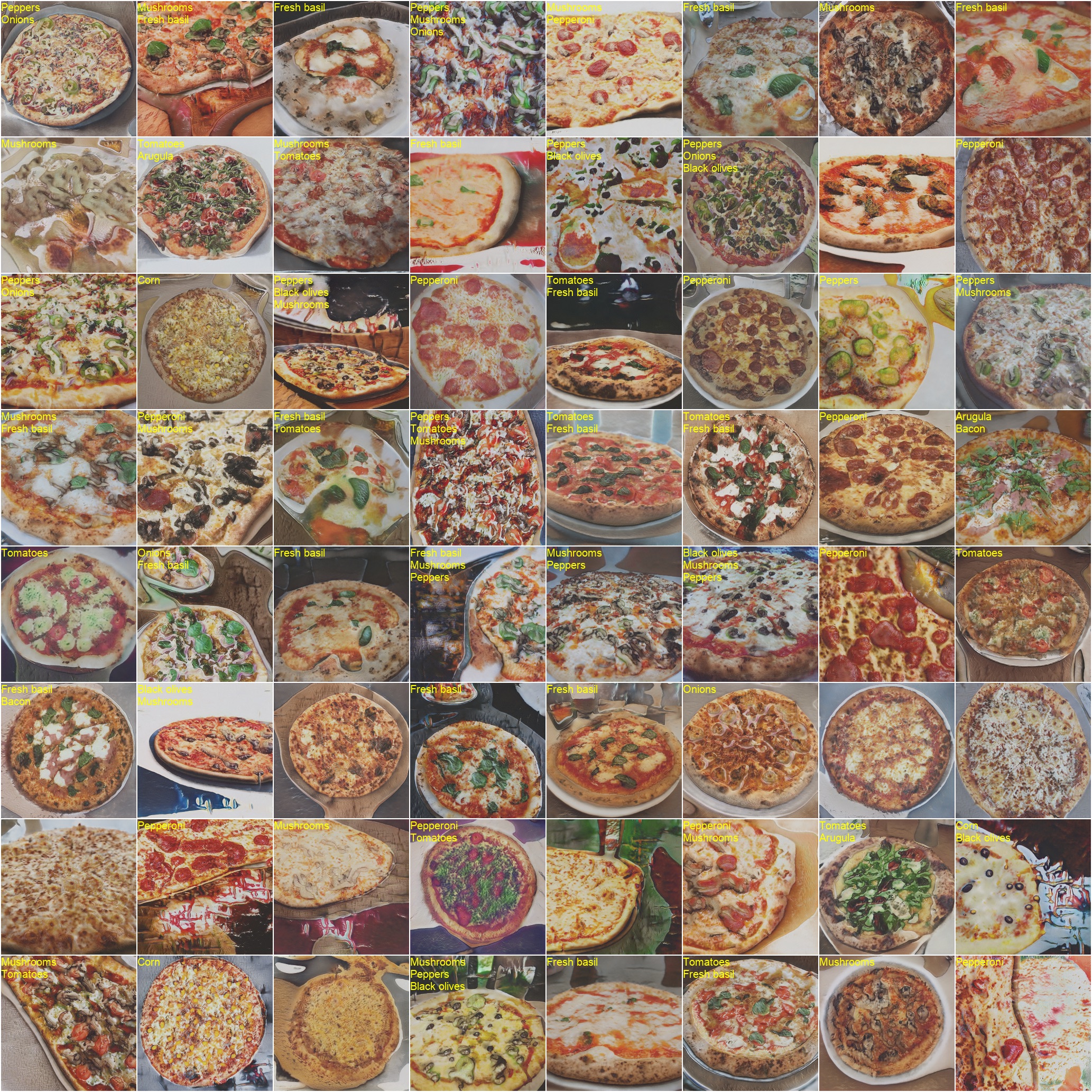}
    \caption{Images generated using our \MPG, ingredients and style noise are randomly sampled. The ingredients are shown on top left corner in each image. The diversity level (\eg truncation) is set to be $1.00$. Please check~\cite{karras2019style} for how to accomplish this using the truncation trick.}
    \label{fig:cs_1.00}
\end{figure*}

\begin{figure*}[ht]
    \centering
    \includegraphics[width=\textwidth]{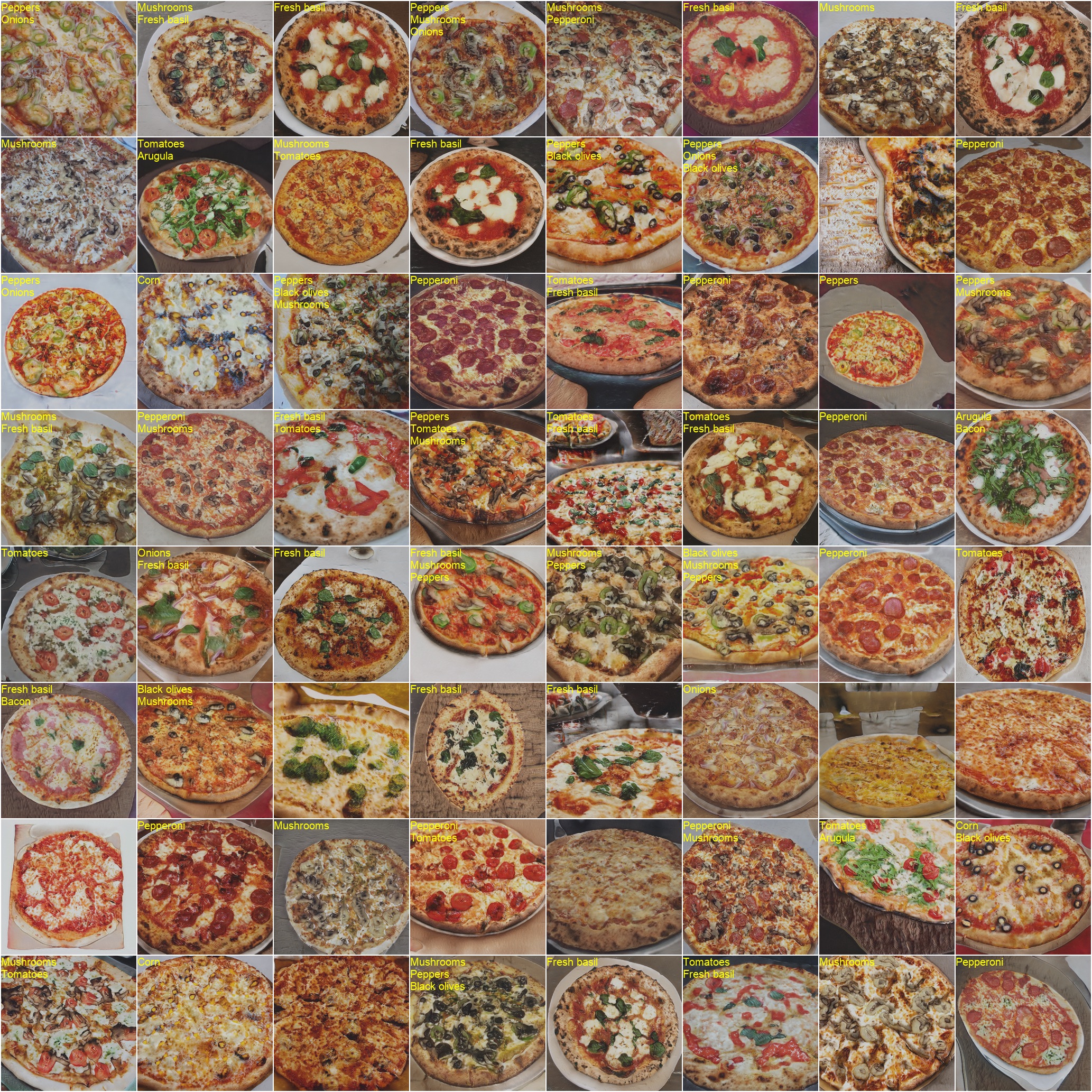}
    \caption{Images generated using our \MPG, ingredients and style noise are randomly sampled. The ingredients are shown on top left corner in each image. The diversity level (\eg truncation) is set to be $0.75$. Please check~\cite{karras2019style} for how to accomplish this using the truncation trick.}
    \label{fig:cs_0.75}
\end{figure*}

\begin{figure*}[ht]
    \centering
    \includegraphics[width=\textwidth]{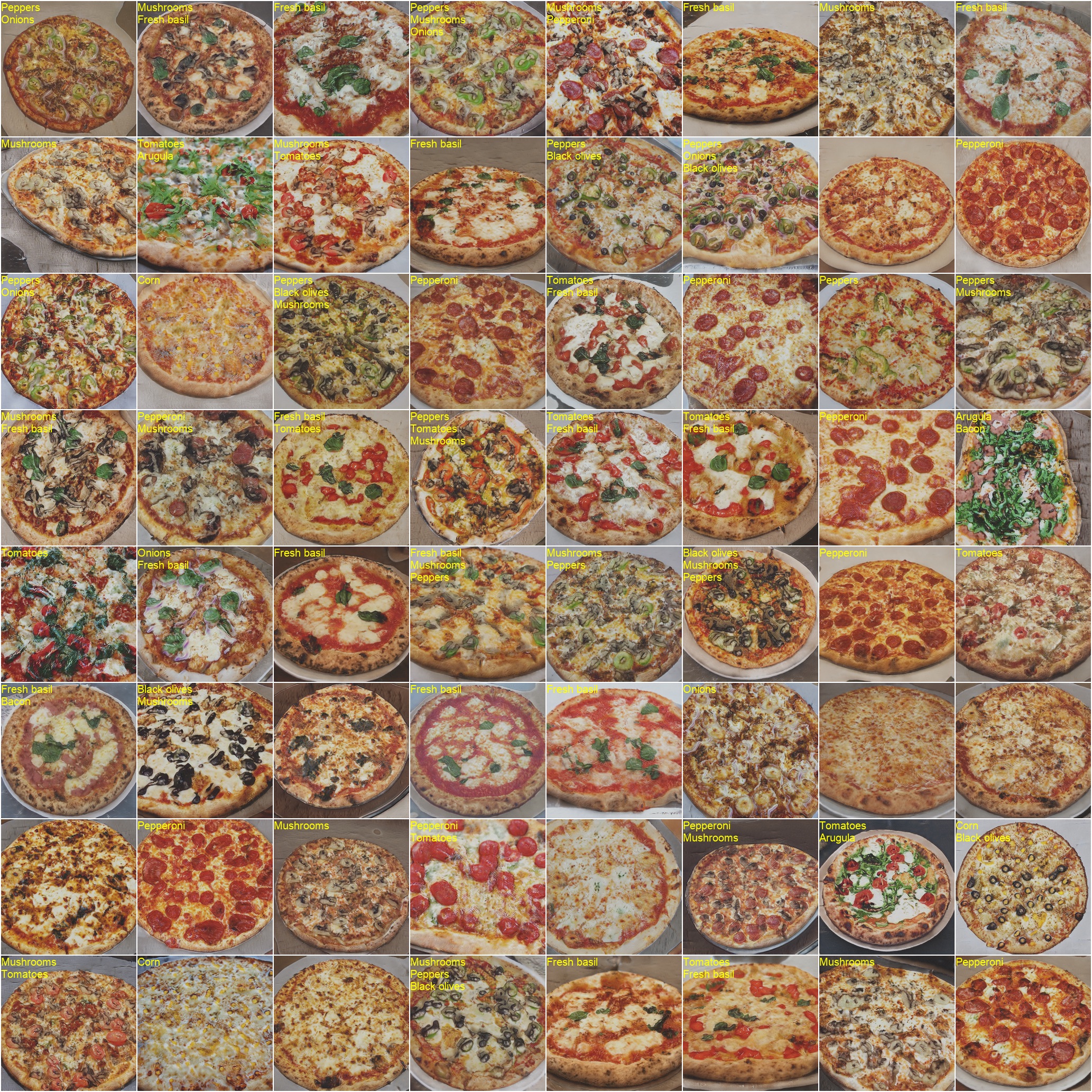}
    \caption{Images generated using our \MPG, ingredients and style noise are randomly sampled. The ingredients are shown on top left corner in each image. The diversity level (\eg truncation) is set to be $0.50$. Please check~\cite{karras2019style} for how to accomplish this using the truncation trick.}
    \label{fig:cs_0.50}
\end{figure*}

\begin{figure*}[ht]
    \centering
    \includegraphics[width=\textwidth]{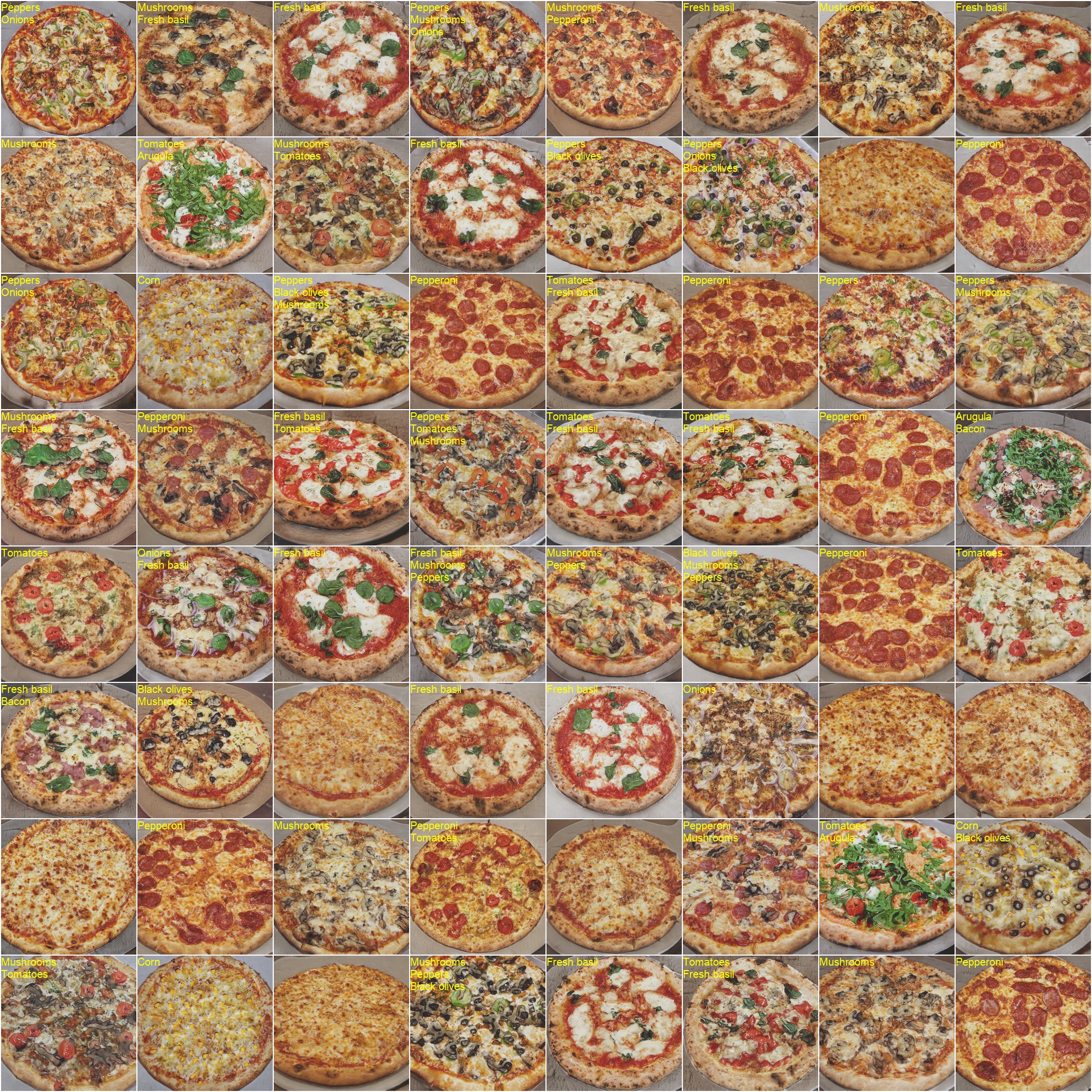}
    \caption{Images generated using our \MPG, ingredients and style noise are randomly sampled. The ingredients are shown on top left corner in each image. The diversity level (\eg truncation) is set to be $0.25$. Please check~\cite{karras2019style} for how to accomplish this using the truncation trick.}
    \label{fig:cs_0.25}
\end{figure*}

\begin{figure*}[ht]
    \centering
    \includegraphics[width=\textwidth]{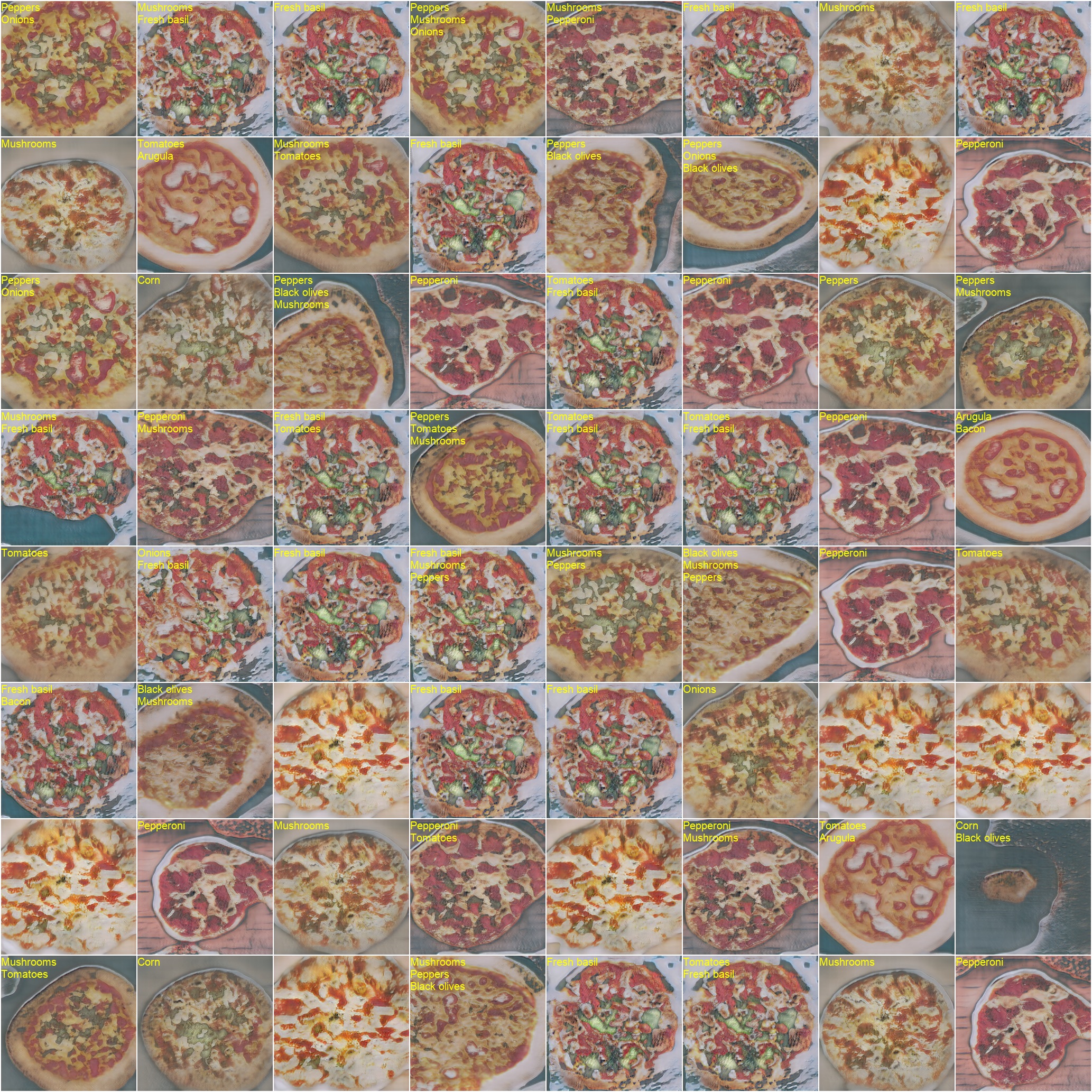}
    \caption{Images generated using StackGAN2~\cite{zhang2018stackgan++}, ingredients and style noise are randomly sampled. The ingredients are shown on top left corner in each image.}
    \label{fig:stackgan2}
\end{figure*}

\begin{figure*}[ht]
    \centering
    \includegraphics[width=\textwidth]{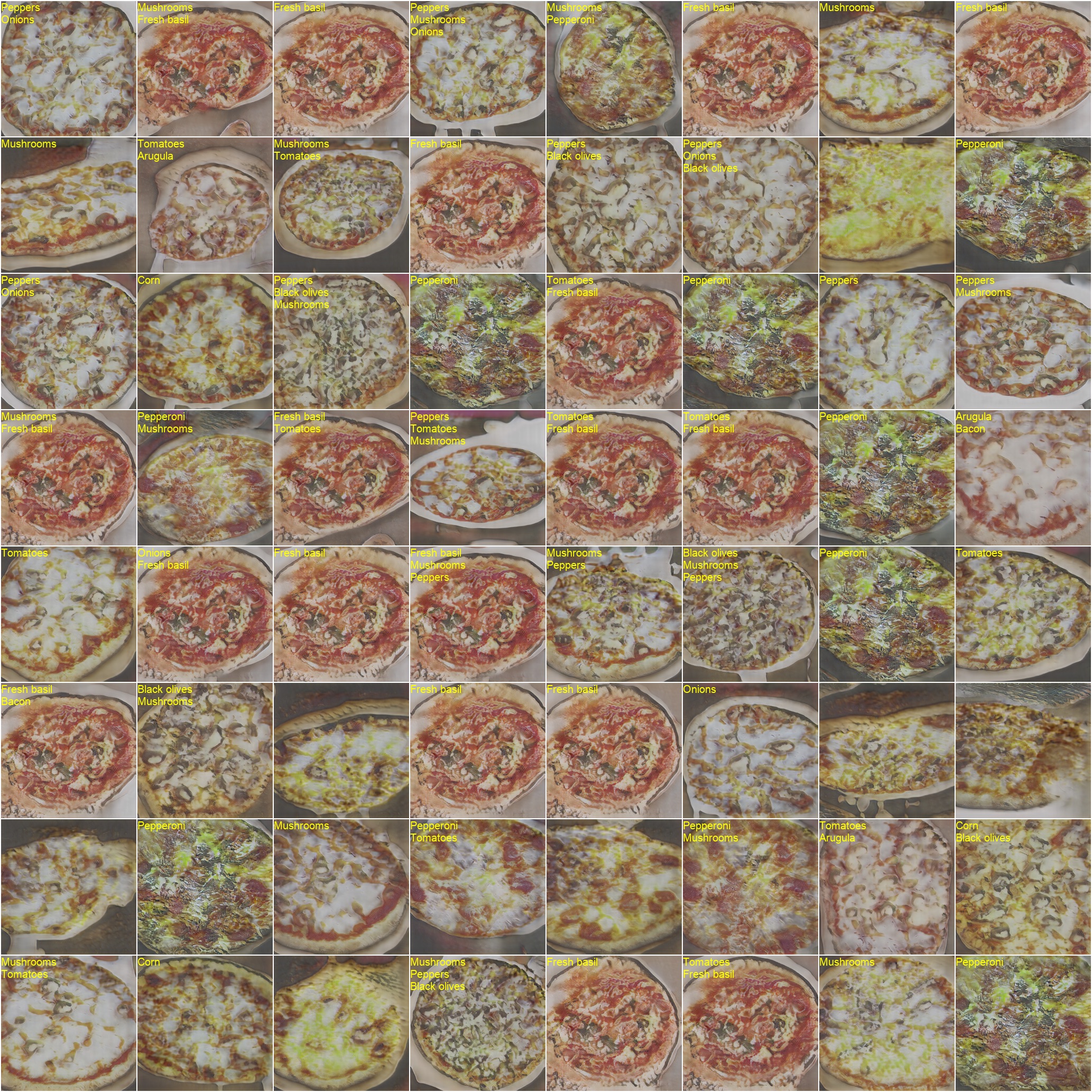}
    \caption{Images generated using CookGAN~\cite{han2020cookgan}, ingredients and style noise are randomly sampled. The ingredients are shown on top left corner in each image.}
    \label{fig:cookgan}
\end{figure*}

\begin{figure*}[ht]
    \centering
    \includegraphics[width=\textwidth]{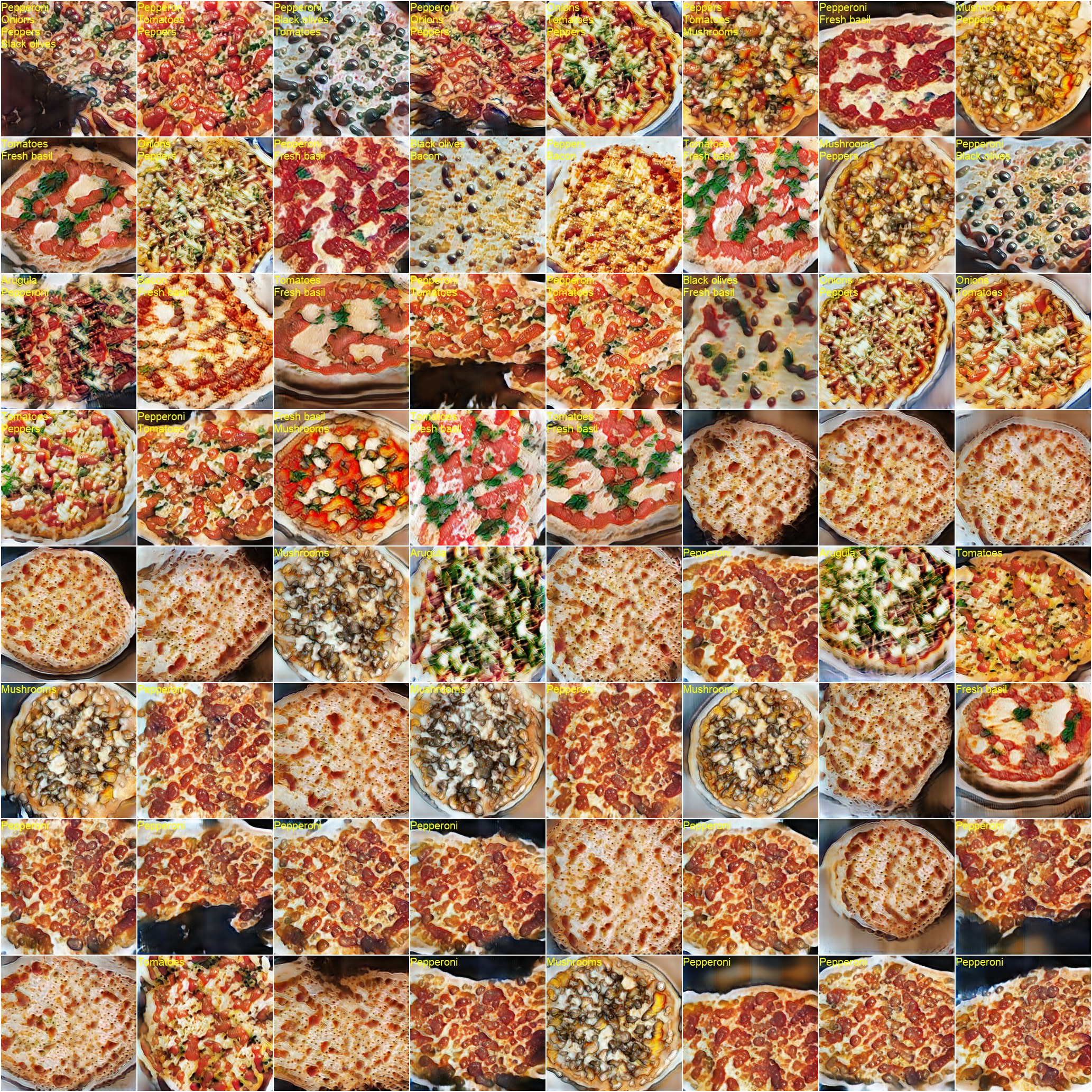}
    \caption{Images generated using AttnGAN~\cite{xu2018attngan}, ingredients and style noise are randomly sampled. The ingredients are shown on top left corner in each image.}
    \label{fig:attngan}
\end{figure*}

We provide additional samples from our proposed \MPG and other state-of-the-art models in Figures~\ref{fig:cs_1.00} through~\ref{fig:attngan}.  They illustrate:

\begin{itemize}
    \item \autoref{fig:cs_1.00} - \autoref{fig:cs_0.25}: Example images generated using our \MPG under different diversity levels.
    \item \autoref{fig:stackgan2}: Example images generated using StackGAN2~\cite{zhang2018stackgan++}.
    \item \autoref{fig:cookgan}: Example images generated using CookGAN~\cite{han2020cookgan}.
    \item \autoref{fig:attngan}: Example images generated using AttnGAN~\cite{xu2018attngan}.
\end{itemize}

\end{document}